\documentclass[11pt]{article}

\usepackage[final]{acl}
\usepackage{lineno}
\usepackage{times}
\usepackage{booktabs}
\usepackage{latexsym}
\usepackage{amsmath}

\usepackage[T1]{fontenc}

\usepackage[utf8]{inputenc}

\usepackage{microtype}

\usepackage{inconsolata}

\usepackage{graphicx}
\usepackage{amssymb}

\usepackage{xcolor}

\usepackage{enumitem}

\usepackage{xspace}

\newcommand{\audioratio}{\textsc{AudioRatio}\xspace}
\newcommand{\audioconsistency}{\textsc{AudioConsistency}\xspace}
\newcommand{\audioentropy}{\textsc{AudioEntropy}\xspace}
\newcommand{\textentropy}{\textsc{TextEntropy}\xspace}

\newcommand{\voxpopuli}{\textsc{VoxPopuli}\xspace}
\newcommand{\callhome}{\textsc{CALLHOME}\xspace}
\newcommand{\fleurs}{\textsc{Fleurs}\xspace}

\newcommand{\qwen}{\texttt{Qwen-2-Audio}\xspace}
\newcommand{\voxtral}{\texttt{Voxtral-3B}\xspace}

\title{Detecting Hallucinations in SpeechLLMs at Inference Time Using Attention Maps}

\author{
  Jonas Waldendorf$^{1,*}$ \quad
  Bashar Awwad Shiekh Hasan$^{2}$ \quad
  Evgenii Tsymbalov$^{2}$\\
  $^{1}$University of Edinburgh \quad $^{2}$Amazon AGI \\
  \texttt{jonas.waldendorf@ed.ac.uk} \quad \texttt{etsymba@amazon.de} \\
  \small{*Work completed during an internship at Amazon AGI}
}

\begin{document}
\maketitle
\begin{abstract}
Hallucinations in Speech Large Language Models (SpeechLLMs) pose significant risks, yet existing detection methods typically rely on gold-standard outputs that are costly or impractical to obtain. Moreover, hallucination detection methods developed for text-based LLMs do not directly capture audio-specific signals. We investigate four attention-derived metrics: \textsc{AudioRatio}, \textsc{AudioConsistency}, \textsc{AudioEntropy}, and \textsc{TextEntropy}, designed to capture pathological attention patterns associated with hallucination, and train lightweight logistic regression classifiers on these features for efficient inference-time detection.

Across automatic speech recognition and speech-to-text translation tasks, evaluations on Qwen-2-Audio and Voxtral-3B show that our approach outperforms uncertainty-based and prior attention-based baselines on in-domain data, achieving improvements of up to +0.23 PR-AUC. We further find that strong performance can be achieved with approximately 100 attention heads, improving out-of-domain generalisation compared to using all heads. While effectiveness is model-dependent and task-specific training is required, our results demonstrate that attention patterns provide a valuable tool for hallucination detection in SpeechLLMs.
\end{abstract}

\section{Introduction}

Speech is a vital modality for many modern technologies, including personal voice assistants~\citep{hoy2018alexa}, automatic transcription services~\citep{radford2023robust}, and audio translation systems~\citep{barrault2023seamlessm4t}. Despite recent advances, speech recognition and translation models can produce hallucinated content, fluent and plausible outputs that are not grounded in the input audio. While typical transcription errors often preserve semantic meaning and may still be interpretable by downstream users, hallucinations can introduce fabricated or misleading information, leading to critical failures. Consequently, identifying hallucinations and models that are predisposed to hallucinate remains an important area of research.

Hallucinations are pathological generations that are not grounded in the input and instead rely on distributional patterns learned from training data. Such outputs are often fluent, yet contain fabrications that alter the semantic content relative to the source audio~\citep{koudounas2025hallucinationbenchmarkspeechfoundation,atwany-etal-2025-lost}. Hallucination detection has been extensively studied in the text modality, for example, in retrieval-augmented generation~\citep{shuster2021retrieval} and machine translation~\citep{guerreiro-etal-2023-hallucinations}. In contrast, existing approaches for SpeechLLMs primarily compare model hypotheses against gold-standard outputs to identify hallucinations~\citep{frieske2024hallucinationsneuralautomaticspeech,koudounas2025hallucinationbenchmarkspeechfoundation}. While effective, such supervised methods require costly data annotation and often depend on external models to perform the comparison.

Lightweight detection methods that leverage the internal representations of SpeechLLMs offer several advantages. First, they can be deployed at inference time for online filtering, preventing harmful outputs at the source, for example, by rerouting problematic inputs. Second, they enable low-cost flagging of pathological generations for offline analysis. Third, they can be combined with complementary signals such as uncertainty estimation (UE) metrics to capture a broader range of errors. Motivated by these benefits, we propose training lightweight classifiers on internal SpeechLLM representations to detect hallucinations at inference time.

Our approach exploits patterns in attention heads as signals for hallucination detection. The underlying intuition is that when models generate outputs that are not grounded in the input, their attention exhibits distinctive patterns that can be detected automatically. Recent work has demonstrated the effectiveness of attention-based signals for hallucination detection in text LLMs~\citep{chuang-etal-2024-lookback,vazhentsev2025uncertaintyawareattentionheadsefficient,NEURIPS2024_3c1e1fdf}. However, these methods have not been adapted to SpeechLLMs, where the input modality introduces fundamentally different attention dynamics. Audio representations are substantially longer than text, and the alignment between input frames and output tokens differs from text-to-text generation.

We address this gap by analysing attention patterns in two SpeechLLMs, \qwen~\citep{chu2024qwen2audiotechnicalreport} and \voxtral~\citep{liu2025voxtral}, and training logistic regression classifiers on audio-specific attention features for hallucination detection.

\paragraph{Contributions}
\begin{itemize}
\item We propose four audio-focused attention metrics, \audioratio, \audioconsistency, \audioentropy, and \textentropy, designed to capture hallucination-related attention patterns in SpeechLLMs.

\item We develop lightweight logistic regression classifiers trained on these attention-derived features. Our models outperform uncertainty estimation methods and existing attention-based baselines for hallucination detection, achieving improvements of up to +0.23 PR-AUC on in-domain ASR data.

\item We show that strong detection performance can be achieved using approximately 100 attention heads. This improves out-of-domain generalisation compared to using all heads, and we further demonstrate that effectiveness varies across models and tasks.
\end{itemize}

\section{Related Work}

\paragraph{Uncertainty Estimation.}
A common approach to hallucination detection relies on UE metrics. Metrics derived from the LLM logits provide a low-cost and effective signal for identifying when a model is uncertain in its generation, which in turn correlates strongly with hallucinated content~\citep{huang2024surveyuncertaintyestimationllms,vashurin-etal-2025-benchmarking,vazhentsev2025uncertaintyawareattentionheadsefficient}. Metrics such as \textsc{Seq-LogProb} (the log probability of the entire output sequence), \textsc{perplexity}, and \textsc{mean entropy} (defined as the average token-level entropy) have all been shown to be effective for hallucination detection~\citep{Malinin2021UncertaintyEI,guerreiro-etal-2023-looking}.

\paragraph{Hallucination Detection via Attention.}
Several approaches exploit attention patterns to detect hallucinations in text LLMs. \citet{vazhentsev2025uncertaintyawareattentionheadsefficient} analyse causal attention to the previous token, while \citet{sriramanan2024llmcheck} average attention maps across layers. \textsc{Lookback-Lens}~\citep{chuang-etal-2024-lookback} train a classifier on the ratio of attention allocated to the input versus the auto-regressive prefix. We build on this line of work by adapting attention-based detection to SpeechLLMs using four audio-focused attention metrics.

Related work in machine translation has shown that hallucinations correlate with reduced diagonal entropy in attention maps~\citep{voita-etal-2021-analyzing,raunak-etal-2021-curious}. We incorporate entropy-based features in our approach, but compute them specifically over audio attention, reflecting the substantial length disparity between audio frame sequences and output tokens.

\paragraph{Reference-Based Methods.}
An alternative class of approaches relies on gold-standard outputs to detect hallucinations. \citet{frieske2024hallucinationsneuralautomaticspeech} study hallucinations in ASR systems and identify them using a combination of semantic similarity between hypotheses and references, together with output fluency under a language model. \citet{koudounas2025hallucinationbenchmarkspeechfoundation} introduce SHALLOW, a benchmark that categorises hallucinations into four types: lexical fabrications, phonetic fabrications, morphological hallucinations, and semantic hallucinations. While such reference-based methods enable fine-grained analysis, they require access to reference transcriptions that are often unavailable in deployment settings. This limitation motivates our reference-free detection approach.

\begin{figure*}[h]
    \centering
    \includegraphics[width=\textwidth
    ]{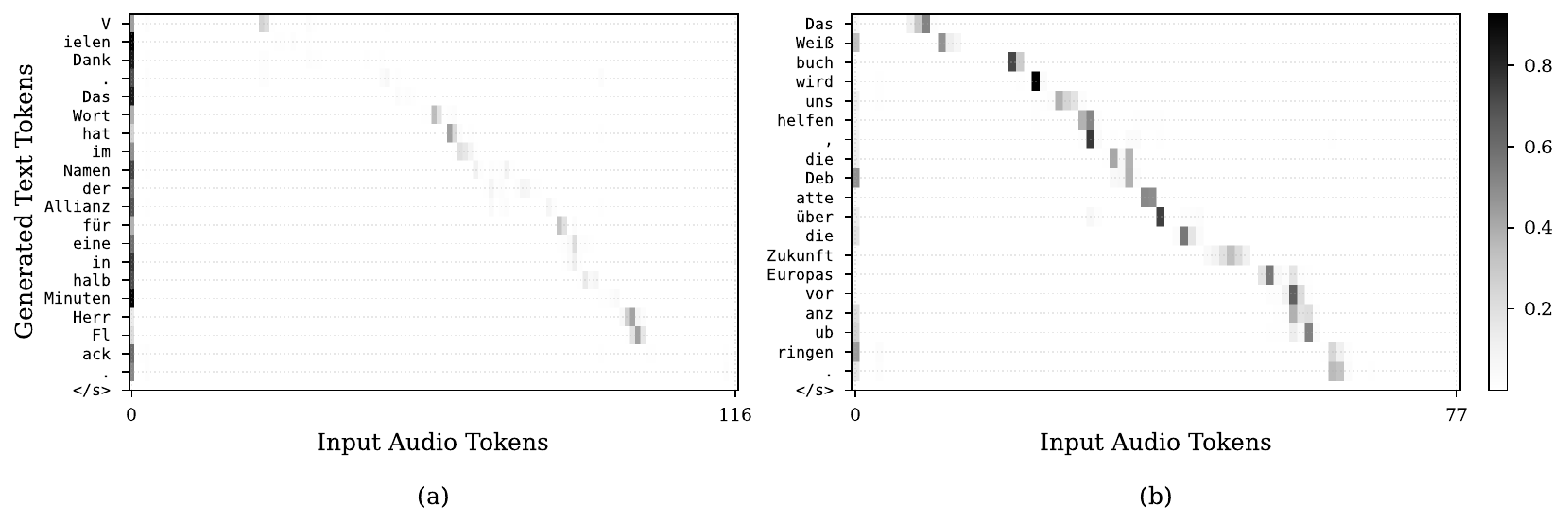}
    \caption{Attention to audio tokens for Layer 25, Head 30 in \voxtral. (a) Hallucination: attention collapses to early audio frames, losing temporal alignment. (b) Correct transcription: diagonal pattern reflects alignment between audio input and generated text.}
    \label{fig:attention}
\end{figure*}

\section{Methodology}

\paragraph{Attention to Input in SpeechLLMs.}
Based on recent findings~\citep{vazhentsev2025uncertaintyawareattentionheadsefficient, Wang2025CalmWhisperRW}, we hypothesise that certain attention heads exhibit distinctive patterns whilst generating hallucinated content, which can be exploited for detection at inference time. We focus on two such patterns. The first consists of diagonal attention structures that encode a temporal relationship between the audio input and the generated text. These patterns often degrade when the model is uncertain, with attention falling back to the early portion of the audio input. The second pattern involves heads that balance attention between the audio input and the auto-regressive text prefix.

Figure~\ref{fig:attention} shows two attention maps from Layer~25, Head~30 of \voxtral, comparing a hallucinated transcription with a correct one. The figure illustrates two effects. First, the diagonal attention pattern, which captures the alignment between the audio input and the generated text, degrades during hallucination. Second, attention falls back to the beginning of the audio input rather than shifting toward the text prompt or the auto-regressive prefix.

\subsection{Metrics}

We experiment with four metrics designed to capture characteristic features of attention heads. Consider a SpeechLLM with $L$ layers and $H$ heads, an input sequence
\[
X = \{x_{a,1}, x_{a,2}, \ldots, x_{a,N} \mid x_{t,1}, x_{t,2}, \ldots, x_{t,M}\}
\]
where $x_{a,i}$ denotes an audio input token and $x_{t,j}$ denotes a text input token, and an output sequence
\[
Y = \{y_1, y_2, \ldots, y_{t-1}\}
\]
We define each metric at a given decoding step $t$ and then describe a shared aggregation strategy.

\begin{description}[style=unboxed,leftmargin=0cm,font=\normalfont\bfseries]

\item[\audioratio:]
The ratio of attention allocated to audio input tokens versus auto-regressive text tokens.

For \audioratio at decoding step $t$ and head $(l,h)$, where ART denotes the auto-regressive text prefix, we compute
\begin{equation}
A^{l,h}_t(\text{Audio}) = \sum_{i=1}^{N} a^{l,h}_{t,i},
\end{equation}

\begin{equation}
A^{l,h}_t(\text{ART}) = \sum_{j=N+M+1}^{N+M+t-1} a^{l,h}_{t,j}.
\end{equation}
The audio ratio is then defined as
\begin{equation}
AR^{l,h}_t =
\frac{A^{l,h}_t(\text{Audio})}
{A^{l,h}_t(\text{Audio}) + A^{l,h}_t(\text{ART})}.
\end{equation}

This metric builds on \textsc{Lookback-Lens} by computing the ratio of attention allocated to input and output tokens, but restricts the input side to audio tokens only, since the text prompt in our tasks contains only instructions.

\item[\audioconsistency:]
The Pearson correlation coefficient between audio attention vectors at consecutive decoding steps. This metric is defined only after the first generation step and is computed as
\begin{equation}
AC^{l,h}_t = r\left(a^{l,h,t}_{1:N}, a^{l,h,t-1}_{1:N}\right)
\end{equation}

As illustrated in Figure~\ref{fig:attention}, this metric aims to capture attention fallback behaviour. When hallucinating, the model often focuses attention at the beginning of the audio input, thereby increasing the similarity between consecutive attention distributions. An important limitation of \audioconsistency is that it explicitly targets heads with diagonal attention patterns. Some heads may attend strongly to early audio positions even during correct generations.

Both entropy-based metrics are computed by selecting the relevant attention weights at decoding step $t$, re-normalising them, and computing entropy as
\begin{equation}
AE^{l,h}_t =
H\left(
\frac{a^{l,h,t}_{1:N}}
{\sum_{i=1}^{N} a^{l,h,t}_{i}}
\right).
\end{equation}

\item[\audioentropy:]
The entropy of re-normalised attention weights over audio input tokens. \audioentropy aims to capture uncertainty in the audio input and provides a useful signal even for heads that do not exhibit clear diagonal attention patterns.

\item[\textentropy:]
The entropy of re-normalised attention weights over text input tokens. The goal of this metric is to capture uncertainty in text-focused attention heads.

\end{description}

For each metric, values are computed at every decoding step and then averaged across all time steps to obtain a single value per layer and head. These values are concatenated into a feature vector of dimension $L \times H$ for each metric. We then use these vectors as input features
to train logistic regression models as hallucination detectors.

\section{Experimental Setup}

We evaluate our models across two speech-related tasks: ASR (automatic speech recognition) and S2TT (speech-to-text translation).

\subsection{Evaluation Datasets}

For ASR, we evaluate on two datasets: \voxpopuli\footnote{\url{https://huggingface.co/datasets/facebook/voxpopuli}} \citet{wang-etal-2021-voxpopuli} and \callhome\footnote{\url{https://catalog.ldc.upenn.edu/LDC97S42}} \citet{canavan1997callhome}. For \voxpopuli, we experiment with the English, German, French, and Spanish language splits. When evaluating hallucination detection, we combine all \voxpopuli test sets for a total of 7,080 sentences. We pre-process \callhome by removing sentences shorter than two words (tokenising on whitespace), leaving 3,916 examples.

For S2TT, we evaluate on the English, German, French, and Spanish subsets of the \fleurs\footnote{\url{https://huggingface.co/datasets/google/fleurs}} \citet{fleurs2022arxiv} multilingual speech translation dataset, totalling 4,613 examples.

\subsection{Labelling Hallucinations}\label{sec:labels}

Training and evaluating hallucination detection models requires binary labels. Since manual annotation is expensive, we first collect a small set of human annotations to calibrate automatic labelling thresholds,  and then apply these thresholds to label the remaining data automatically.

\paragraph{Human Annotation.}
We manually annotated 1,950 examples sampled from the English and German \voxpopuli development sets, identifying 142 hallucinated outputs. An output was labelled as a hallucination if it contained fluent but fabricated content that was not grounded in the input audio.

\begin{table}[htbp]
\centering
\resizebox{\columnwidth}{!}{%
\begin{tabular}{@{}llccc@{}}
\toprule
Dataset & Language & WER & SHS & Hal.\% \\
\midrule
\voxpopuli & De & 13.09 & 11.47 & 5.7 \\
 & En & 7.01 & 8.29 & 1.1 \\
 & Es & 9.47 & 11.61 & 2.6 \\
 & Fr & 10.98 & 10.73 & 3.6 \\
\midrule
\callhome & En & 20.90 & 19.03 & 20.6 \\
\bottomrule
\end{tabular}%
}
\caption{WER, SHS, and hallucination percentage for \qwen on \voxpopuli and \callhome test sets.}
\label{tab:hallucination_qwen}
\end{table}

\paragraph{Automatic Labelling.} To label hallucinations automatically, we threshold on a combination of lexical and semantic information. This approach follows prior findings that semantic content is critical for identifying ASR hallucinations \citet{frieske2024hallucinationsneuralautomaticspeech}. We use WER (word error rate) to capture surface-level errors and the \textsc{Semantic Hallucination Score} (SHS) from the SHALLOW benchmark~\citep{koudounas2025hallucinationbenchmarkspeechfoundation} to capture semantic divergence (Appendix~\ref{sec:shs}):

\begin{equation}
\text{Hallucination} = \mathbb{I}[\text{WER} + \text{SHS} > 0.7],
\label{eq:thresh}
\end{equation}
where $\mathbb{I}[\cdot]$ denotes the indicator function.

\paragraph{Threshold Selection.}
The threshold was tuned using stratified five-fold cross-validation on the human-annotated subset. We prioritised high precision (0.979) in order to obtain a clean training signal, accepting lower recall (0.443) as a trade-off.

\begin{table}[h]
\centering
\resizebox{\columnwidth}{!}{%
\begin{tabular}{@{}llccc@{}}
\toprule
Dataset & Language & WER & SHS & Hal.\% \\
\midrule
\voxpopuli & De & 11.66 & 11.59 & 6.3 \\
 & En & 7.16 & 8.76 & 1.1 \\
 & Es & 8.59 & 11.27 & 2.5 \\
 & Fr & 10.46 & 10.60 & 3.3 \\
\midrule
\callhome & En & 18.72 & 16.52 & 15.8 \\
\bottomrule
\end{tabular}%
}
\caption{WER, SHS, and hallucination percentage for \voxtral on \voxpopuli and \callhome test sets.}
\label{tab:hallucination_voxtral}
\end{table}

Tables~\ref{tab:hallucination_qwen} and~\ref{tab:hallucination_voxtral} report hallucination rates for both models across evaluation datasets. Hallucination rates are consistently low on \voxpopuli test sets, but increase substantially, up to 20\%, on the noisier \callhome dataset. Both models exhibit similar trends, although \voxtral generally achieves lower error rates, with the exception of German.

\subsection{Training Data}
We train all logistic regression models on 40,000 examples from the \voxpopuli training data: 10,000 each for English, German, Spanish, and French. This language mix exposes the model to varying hallucination rates across languages (Tables~\ref{tab:hallucination_qwen} and~\ref{tab:hallucination_voxtral}), improving robustness to distribution shifts. Under this setup, \qwen produces 1,537 hallucinations (3.8\%), while \voxtral produces 1,178 hallucinations (2.9\%).

For the S2TT task, we train on the \fleurs training set (16,776 examples), using hallucination labels derived from COMET scores, and evaluate on the held-out test set.

\subsection{Logistic Regression Model Training}
Using the attention-derived features described above, we train logistic regression models for hallucination detection. Model hyperparameters are reported in the Appendix (Table~\ref{tab:hyperparams}). We apply MinMax scaling to \audioentropy and \textentropy to ensure that all feature values lie within the range $[0,1]$.

We employ two feature selection strategies. First, we train an L2-regularised model and rank attention heads by the magnitude of their coefficients, scaled by the original feature standard deviation. Second, we train an L1-regularised model to perform feature pruning. Specifically, we run five-fold cross-validation and retain heads with non-zero coefficients in at least four of the five folds. We then retrain an L2-regularised model using only these retained heads. We refer to this variant as the \textit{Stable Features} model.

\subsection{Evaluation Metrics}

\paragraph{ASR Evaluation.} For ASR, we report F1-score, precision, recall, and PR-AUC, computed using the predicted probabilities from the logistic regression models.

In addition, we report the Prediction Rejection Ratio (PRR)~\citep{malinin-etal-2017-incorporating, Malinin2021UncertaintyEI}, which measures the effectiveness of predicted probabilities in rejecting low-quality samples. Intuitively, PRR quantifies how closely probability-based rejection approaches oracle performance, where $\text{PRR}=1$ corresponds to a perfect ordering with respect to a quality metric. We report PRR at 10\% rejection for \voxpopuli and at 30\% rejection for \callhome, using SHS as the quality metric. These rejection rates reflect the approximate prevalence of hallucinations in each dataset. Further details are provided in Appendix~\ref{sec:prr}.

\paragraph{S2TT Evaluation.}
For S2TT, we use XCOMET-XL\footnote{\url{https://huggingface.co/Unbabel/XCOMET-XL}}~\citep{guerreiro-etal-2024-xcomet}, which has been shown to correlate strongly with hallucinations in machine translation. Based on the empirical distribution of COMET scores, we label the bottom 5\% of examples as hallucinations. We report F1-score, precision, recall, and PR-AUC, following the same protocol as for ASR. We also report PRR@10\%, using COMET as the quality metric.

\subsection{Baselines}

As baselines, we consider simple UE metrics, namely \textsc{Mean Entropy} (the mean token-level entropy) and \textsc{Perplexity}, which have been shown to correlate with hallucinations in text summarisation and machine translation. In addition, we include two attention-based baselines, \textsc{RAUQ (Entropy)} and \textsc{AttentionScore}, as attention-based methods for hallucination detection.

\section{Results}

\subsection{ASR Results}

Tables~\ref{tab:hallucination_detection_qwen} and~\ref{tab:hallucination_detection_voxtral} report hallucination detection performance for \qwen and \voxtral, respectively, on the \voxpopuli and \callhome test sets. We present results for logistic regression models trained using the following feature configurations.

\textbf{Combined} concatenates all four attention metrics across all layers and heads, yielding $L \times H$ features per metric.
\textbf{\audioratio Only} uses only the \audioratio metric across all layers and heads, which was the best-performing single metric on the \voxpopuli validation set.
\textbf{Top N} selects the top $N$ layer–head pairs per metric based on coefficient magnitude, with $N$ tuned separately for each model on the validation set.

\begin{table}[htbp]
\centering
\resizebox{\columnwidth}{!}{%
\begin{tabular}{@{}llccccccc@{}}
\toprule
Dataset & Method & Hal.\% & Acc & F1 & Prec & Rec & PR-AUC & PRR@k \\
\midrule
\multicolumn{9}{@{}l}{\textsc{Baselines}} \\
\midrule
\multicolumn{9}{@{}l}{\textit{\voxpopuli}} \\
 & Mean Entropy & 3.33 & 0.97 & 0.50 & 0.50 & 0.50 & 0.49 & 0.43 \\
 & Perplexity & 3.18 & 0.97 & 0.50 & 0.51 & 0.49 & 0.49 & 0.43 \\
 & RAUQ Entropy & 3.42 & 0.96 & 0.46 & 0.45 & 0.47 & 0.47 & 0.46 \\
 & Attention Score & 1.23 & 0.96 & 0.08 & 0.15 & 0.06 & 0.04 & 0.30 \\
 & Random & 50.27 & 0.50 & 0.06 & 0.03 & 0.50 & -- & -- \\
\addlinespace
\multicolumn{9}{@{}l}{\textit{\callhome}} \\
 & Mean Entropy & 38.23 & 0.76 & 0.58 & 0.45 & 0.83 & 0.67 & 0.59 \\
 & Perplexity & 37.89 & 0.76 & 0.58 & 0.45 & 0.83 & 0.69 & 0.61 \\
 & RAUQ Entropy & 24.40 & 0.80 & 0.56 & 0.52 & 0.61 & 0.61 & 0.54 \\
 & Attention Score & 0.00 & 0.79 & 0.00 & 0.00 & 0.00 & 0.14 & -0.13 \\
 & Random & 50.03 & 0.50 & 0.29 & 0.20 & 0.49 & -- & -- \\
\midrule
\multicolumn{9}{@{}l}{\textsc{Logistic Regression}} \\
\midrule
\multicolumn{9}{@{}l}{\textit{Combined (4096 features)}} \\
\voxpopuli & LR & 3.62 & 0.97 & 0.55 & 0.52 & 0.57 & 0.58 & 0.51 \\
\callhome & LR & 75.99 & 0.43 & 0.41 & 0.26 & 0.96 & 0.61 & 0.53 \\
\addlinespace
\multicolumn{9}{@{}l}{\textit{\audioratio (1024 features)}} \\
\voxpopuli & LR & 3.57 & 0.97 & 0.56 & 0.54 & 0.58 & 0.56 & 0.49 \\
\callhome & LR & 66.97 & 0.52 & 0.45 & 0.29 & 0.95 & 0.60 & 0.53 \\
\addlinespace
\multicolumn{9}{@{}l}{\textit{Top 75 (300 features)}} \\
\voxpopuli & LR & 3.87 & 0.97 & 0.52 & 0.49 & 0.57 & 0.58 & 0.49 \\
\callhome & LR & 79.05 & 0.40 & 0.40 & 0.25 & 0.97 & 0.59 & 0.50 \\
\bottomrule
\end{tabular}%
}
\caption{Hallucination detection performance (F1, Precision, Recall, PR-AUC, and PRR@$k$) for \qwen on \voxpopuli and \callhome test sets, with $k{=}10\%$ and $k{=}30\%$ respectively.}
\label{tab:hallucination_detection_qwen}
\end{table}

\paragraph{Attention-based features outperform baselines on in-domain data.}
On the \voxpopuli test set, logistic regression models consistently outperform uncertainty-based baselines for both SpeechLLMs. For \qwen, the strongest baseline, \textsc{Mean Entropy}, achieves an F1-score of 0.50 and a PR-AUC of 0.49, whereas logistic regression attains up to 0.56 F1 and 0.58 PR-AUC. The improvement is substantially larger for \voxtral. Here, \textsc{Mean Entropy} reaches 0.42 F1 and 0.44 PR-AUC, while \audioratio achieves 0.64 F1 and 0.67 PR-AUC, corresponding to a gain of 0.23 PR-AUC.

Label-free evaluation supports these results. PRR@k improves by 0.05 for \qwen and by 0.13 for \voxtral relative to the strongest baseline that does not use attention. Together, these findings demonstrate that attention-based features provide a strong discriminative signal for hallucination detection on in-domain data. Qualitative examples are provided in Appendix~\ref{sec:examples}.

\begin{table}[htbp]
\centering
\resizebox{\columnwidth}{!}{%
\begin{tabular}{@{}llccccccc@{}}
\toprule
Dataset & Method & Hal.\% & Acc & F1 & Prec & Rec & PR-AUC & PRR@k \\
\midrule
\multicolumn{9}{@{}l}{\textsc{Baselines}} \\
\midrule
\multicolumn{9}{@{}l}{\textit{\voxpopuli}} \\
 & Mean Entropy & 2.10 & 0.97 & 0.42 & 0.55 & 0.34 & 0.44 & 0.43 \\
 & Perplexity & 3.12 & 0.96 & 0.40 & 0.42 & 0.39 & 0.41 & 0.40 \\
 & RAUQ Entropy & 2.03 & 0.96 & 0.35 & 0.47 & 0.28 & 0.32 & 0.43 \\
 & Attention Score & 7.12 & 0.91 & 0.17 & 0.12 & 0.26 & 0.09 & 0.10 \\
 & Random & 50.27 & 0.50 & 0.06 & 0.03 & 0.49 & -- & -- \\
\addlinespace
\multicolumn{9}{@{}l}{\textit{\callhome}} \\
 & Mean Entropy & 15.83 & 0.86 & 0.55 & 0.55 & 0.55 & 0.59 & 0.57 \\
 & Perplexity & 25.18 & 0.79 & 0.50 & 0.40 & 0.64 & 0.56 & 0.54 \\
 & RAUQ Entropy & 11.56 & 0.86 & 0.51 & 0.60 & 0.44 & 0.52 & 0.53 \\
 & Attention Score & 53.92 & 0.46 & 0.23 & 0.15 & 0.51 & 0.15 & -0.01 \\
 & Random & 50.03 & 0.49 & 0.22 & 0.15 & 0.46 & -- & -- \\
\midrule
\multicolumn{9}{@{}l}{\textsc{Logistic Regression}} \\
\midrule
\multicolumn{9}{@{}l}{\textit{Combined (3840 features)}} \\
\voxpopuli & LR & 3.23 & 0.98 & 0.64 & 0.65 & 0.62 & 0.69 & 0.56 \\
\callhome & LR & 32.80 & 0.76 & 0.50 & 0.37 & 0.77 & 0.55 & 0.52 \\
\addlinespace
\multicolumn{9}{@{}l}{\textit{\audioratio (960 features)}} \\
\voxpopuli & LR & 3.04 & 0.98 & 0.64 & 0.68 & 0.61 & 0.67 & 0.54 \\
\callhome & LR & 29.13 & 0.79 & 0.52 & 0.40 & 0.74 & 0.58 & 0.55 \\
\addlinespace
\multicolumn{9}{@{}l}{\textit{Top 75 (300 features)}} \\
\voxpopuli & LR & 2.99 & 0.98 & 0.62 & 0.66 & 0.58 & 0.68 & 0.55 \\
\callhome & LR & 27.26 & 0.80 & 0.55 & 0.43 & 0.74 & 0.61 & 0.57 \\
\bottomrule
\end{tabular}%
}
\caption{Hallucination detection performance (F1, Precision, Recall, PR-AUC, and PRR@$k$) for \voxtral on \voxpopuli and \callhome test sets, with $k{=}10\%$ and $k{=}30\%$ respectively.}
\label{tab:hallucination_detection_voxtral}
\end{table}

\paragraph{Uncertainty baselines excel on noisy speech.}
Baseline performance is notably stronger on \callhome than on \voxpopuli for both models. For \qwen, \textsc{Perplexity} achieves a PR-AUC of 0.69 on \callhome compared to 0.49 on \voxpopuli, while for \voxtral, \textsc{Mean Entropy} achieves 0.59 versus 0.44. We attribute this effect to the inherently noisy nature of \callhome. Conversational speech with overlapping speakers and frequent disfluencies induces higher model uncertainty, making uncertainty-based metrics more effective. This interpretation is supported by the average \textsc{Mean Entropy} values, which are approximately 0.10 for \voxpopuli and 0.30 for \callhome. We observe a consistent pattern across models: \voxtral achieves stronger overall ASR performance, while uncertainty-based methods are comparatively more effective for \qwen. In contrast, our attention-based approach performs best for stronger models, cleaner data, and settings in which hallucinations are relatively rare.

\paragraph{Out-of-domain generalization is model-dependent.}
Logistic regression performance on \callhome differs markedly between the two models. For \qwen, logistic regression predicts a large number of false positives, with predicted hallucination rates ranging from 66.97\% to 79.05\%, compared to an actual rate of 20.90\%. As a result, the best logistic regression model achieves an F1-score of only 0.45 using \audioratio, compared to 0.58 for both \textsc{Perplexity} and \textsc{Mean Entropy}. PRR@k similarly drops by 0.11 relative to the strongest baseline for the Top 75 configuration.

In contrast, \voxtral maintains competitive out-of-domain performance. The Top 75 model achieves an F1-score of 0.55 and a PR-AUC of 0.61 on \callhome, matching or exceeding \textsc{Mean Entropy}. Notably, PRR@k is higher on \callhome than on \voxpopuli for \voxtral in all configurations except the Combined model. This suggests that, for \voxtral, attention-based features generalise well despite the absence of comparable training data. Overall, these results indicate that out-of-domain performance benefits from combining multiple attention head metrics, but remains strongly model-dependent.

\begin{figure}[h]
    \centering
    \includegraphics[width=\columnwidth]{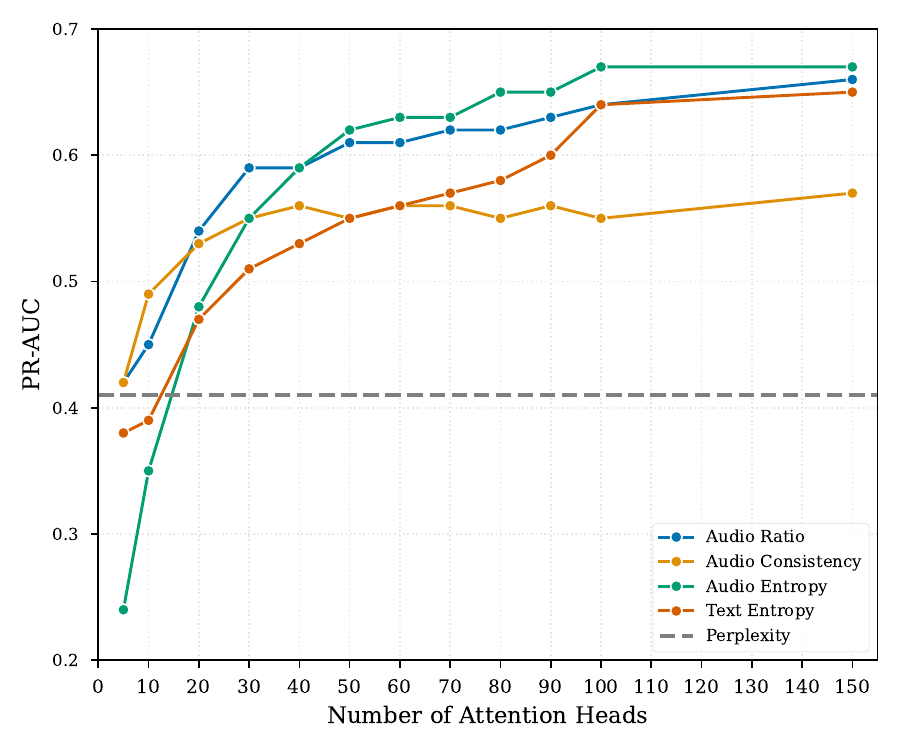}
    \caption{PR-AUC as a function of feature count for \voxtral on \voxpopuli. Each curve represents a different attention metric, with heads ranked independently by coefficient magnitude using L2 logistic regression.}
    \label{fig:scaling}
\end{figure}

\paragraph{\qwen performance gaps reflect threshold sensitivity.}
Despite \qwen’s low F1-scores on \callhome, PR-AUC remains comparable to \voxpopuli, ranging from 0.61 to 0.62 on \callhome versus 0.56 to 0.58 on \voxpopuli. This indicates that the underlying attention features still contain discriminative information. The elevated false positive rate appears to be driven primarily by threshold miscalibration under distribution shift, rather than by fundamentally different attention behaviour. PRR@k results also support that the gap between datasets is substantially smaller than that suggested by the F1-scores. The label-free metric confirms that the proposed approach can still prioritise semantically problematic outputs even when binary decision thresholds are poorly calibrated.

\paragraph{Fewer features improve out-of-domain generalization.}
Figure~\ref{fig:scaling} plots PR-AUC as a function of feature count for \voxtral for \voxpopuli. Performance improvements plateau after approximately 100 attention heads per metric. With as few as five features, all metrics perform similarly to, or worse than, the \textsc{Perplexity} baseline, indicating that while individual heads may carry signal, combining multiple heads is necessary for robust performance. \audioconsistency performs well with relatively few heads but saturates earlier, which is expected given that it targets specific diagonal attention patterns. In contrast, \audioentropy requires aggregating signals from approximately 30 or more heads to reach comparable PR-AUC, after which performance scales more smoothly.

Table~\ref{tab:scaling} further shows that reducing the number of features trades a small amount of in-domain performance for improved generalisation. The \textit{Stable Features} model (see Appendix: Table \ref{tab:stable_features}), which uses 99 features, shows a reduction of 0.02 PR-AUC on \voxpopuli compared to Top 75, which uses 300 features, but achieves the highest PR-AUC of any logistic regression model on \callhome at 0.64. Using only \audioratio features with an equivalent feature budget identifies fewer hallucinations, suggesting that combining metrics becomes increasingly important as the feature count is reduced. Analysis of the selected features shows that the L1-regularised model prioritises \audioratio and \audioconsistency heads, as reported in Appendix~\ref{tab:stable_features}. This observation is consistent with the scaling trends shown in Figure~\ref{fig:scaling}.

\begin{table}[htbp]
\centering
\resizebox{\columnwidth}{!}{%
\begin{tabular}{@{}llccccccc@{}}
\toprule
Dataset & Method & Hal.\% & Acc & F1 & Prec & Rec & PR-AUC & PRR@k \\
\midrule
\multicolumn{9}{@{}l}{\textit{Top 25 (100 features)}} \\
\voxpopuli & LR & 2.98 & 0.97 & 0.60 & 0.64 & 0.56 & 0.65 & 0.53 \\
\callhome & LR & 25.08 & 0.81 & 0.54 & 0.44 & 0.70 & 0.61 & 0.56 \\
\addlinespace
\multicolumn{9}{@{}l}{\textit{\audioratio Only (100 features)}} \\
\voxpopuli & LR & 2.78 & 0.98 & 0.60 & 0.67 & 0.55 & 0.64 & 0.49 \\
\callhome & LR & 21.44 & 0.83 & 0.55 & 0.47 & 0.64 & 0.58 & 0.58 \\
\addlinespace
\multicolumn{9}{@{}l}{\textit{Stable Features (99 features, threshold=0.8)}} \\
\voxpopuli & LR & 2.80 & 0.98 & 0.60 & 0.66 & 0.55 & 0.66 & 0.52 \\
\callhome & LR & 29.96 & 0.79 & 0.53 & 0.41 & 0.77 & 0.64 & 0.58 \\
\bottomrule
\end{tabular}%
}
\caption{Hallucination detection performance (F1, Precision, Recall, PR-AUC, and PRR@$k$) for \voxtral with reduced feature sets on \voxpopuli and \callhome test sets, with $k{=}10\%$ and $k{=}30\%$ respectively.}
\label{tab:scaling}
\end{table}

\subsection{S2TT Results}

\paragraph{Training on ASR data does not generalise to S2TT.}
We first evaluate whether logistic regression models trained on ASR data can transfer to the S2TT task. Using the best-performing ASR configurations, namely Top 75 logistic regression for both \qwen and \voxtral, we observe PR-AUC scores of only 0.15 and 0.08, respectively, which is only marginally above random performance. PRR@k metrics corroborate this result, with values of 0.30 for \qwen and 0.16 for \voxtral, compared to baseline PRR@k scores of 0.46 and 0.44, confirming that attention-based hallucination detectors trained on ASR data do not generalise to S2TT. This raises the question of whether attention patterns differ fundamentally between tasks, or whether the use of different labelling schemes, WER and SHS for ASR versus COMET for S2TT, necessitates task-specific classifiers.

\begin{table}[htbp]
\centering
\resizebox{\columnwidth}{!}{%
\begin{tabular}{@{}lccccccc@{}}
\toprule
Method & Hal.\% & Acc & F1 & Prec & Rec & PR-AUC & PRR@k \\
\midrule
\multicolumn{8}{@{}l}{\textsc{Trained on S2TT Data}} \\
\midrule
\multicolumn{8}{@{}l}{\textit{Baselines}} \\
Mean Entropy & 0.21 & 0.95 & 0.08 & 1.00 & 0.04 & 0.25 & 0.45 \\
Perplexity & 0.23 & 0.95 & 0.08 & 0.90 & 0.04 & 0.23 & 0.39 \\
RAUQ Entropy & 0.18 & 0.95 & 0.07 & 1.00 & 0.04 & 0.25 & 0.46 \\
Attention Score & 14.06 & 0.84 & 0.14 & 0.10 & 0.27 & 0.08 & 0.15 \\
Random & 50.99 & 0.49 & 0.09 & 0.05 & 0.50 & -- & -- \\
\addlinespace
\multicolumn{8}{@{}l}{\textit{Combined (4096 features)}} \\
LR & 1.32 & 0.96 & 0.29 & 0.69 & 0.18 & 0.43 & 0.67 \\
\addlinespace
\multicolumn{8}{@{}l}{\textit{\audioratio (1024 features)}} \\
LR & 1.30 & 0.95 & 0.23 & 0.56 & 0.15 & 0.39 & 0.64 \\
\addlinespace
\multicolumn{8}{@{}l}{\textit{Top 150 (600 features)}} \\
LR & 1.51 & 0.95 & 0.28 & 0.61 & 0.18 & 0.44 & 0.67 \\
\midrule
\multicolumn{8}{@{}l}{\textsc{Trained on ASR Data}} \\
\midrule
\multicolumn{8}{@{}l}{\textit{Top 75 (300 features)}} \\
LR & 89.66 & 0.14 & 0.09 & 0.05 & 0.86 & 0.15 & 0.30 \\
\bottomrule
\end{tabular}%
}
\caption{Hallucination detection performance (F1, Precision, Recall, PR-AUC, and PRR@$k$) for \qwen on the \fleurs S2TT test set, with $k{=}10\%$.}
\label{tab:s2tt_qwen}
\end{table}

\paragraph{Baselines struggle on S2TT.} 
Baseline performance on S2TT is substantially weaker than on ASR. The strongest baseline achieves a PR-AUC of only 0.25 for \qwen and 0.17 for \voxtral, compared to 0.49 and 0.44, respectively, on the \voxpopuli ASR test set. This suggests that uncertainty-based metrics, while effective for ASR, are less suited to detecting hallucinations in the speech-to-text translation setting. This finding is notable, given that such metrics are often strong baselines in text-based machine translation.

\paragraph{In-domain training yields substantial improvements over baselines.} 
Training logistic regression models directly on S2TT data yields substantial performance gains over all baselines. For \qwen, the Top 150 configuration achieves a PR-AUC of 0.44 and an F1-score of 0.28, corresponding to improvements of 0.19 PR-AUC and 0.20 F1 over RAUQ Entropy. For \voxtral, Top 300 reaches a PR-AUC of 0.44 and an F1-score of 0.37, improving on \textsc{Mean Entropy} by 0.27 and 0.26, respectively. Label-free evaluation further supports these results. PRR@k scores reach 0.67 for \qwen and 0.68 for \voxtral, indicating that logistic regression effectively prioritises low-quality translations. However, for both models, recall remains low, suggesting that while attention-based features reliably identify severe hallucinations, they are less sensitive to more subtle S2TT errors.

\begin{table}[htbp]
\centering
\resizebox{\columnwidth}{!}{%
\begin{tabular}{@{}lccccccc@{}}
\toprule
Method & Hal.\% & Acc & F1 & Prec & Rec & PR-AUC & PRR@k \\
\midrule
\multicolumn{8}{@{}l}{\textsc{Trained on S2TT data}} \\
\midrule
\multicolumn{8}{@{}l}{\textit{Baselines}} \\
Mean Entropy & 1.13 & 0.95 & 0.11 & 0.31 & 0.07 & 0.17 & 0.32 \\
Perplexity & 0.17 & 0.95 & 0.05 & 0.75 & 0.03 & 0.15 & 0.27 \\
RAUQ Entropy & 1.15 & 0.95 & 0.13 & 0.34 & 0.08 & 0.16 & 0.35 \\
Attention Score & 83.57 & 0.20 & 0.10 & 0.05 & 0.90 & 0.06 & 0.02 \\
Random & 49.04 & 0.51 & 0.09 & 0.05 & 0.51 & -- & -- \\
\addlinespace
\multicolumn{8}{@{}l}{\textit{Combined (3840 features)}} \\
LR & 2.25 & 0.95 & 0.37 & 0.60 & 0.27 & 0.43 & 0.66 \\
\addlinespace
\multicolumn{8}{@{}l}{\textit{\audioratio (960 features)}} \\
LR & 1.80 & 0.97 & 0.35 & 0.45 & 0.29 & 0.38 & 0.66 \\
\addlinespace
\multicolumn{8}{@{}l}{\textit{Top 150 (600 features)}} \\
LR & 2.51 & 0.95 & 0.37 & 0.55 & 0.28 & 0.44 & 0.68 \\
\midrule
\multicolumn{8}{@{}l}{\textsc{Trained on ASR Data}} \\
\midrule
\multicolumn{8}{@{}l}{\textit{Top 75 (300 features)}} \\
LR & 83.92 & 0.20 & 0.10 & 0.05 & 0.92 & 0.08 & 0.16 \\
\bottomrule
\end{tabular}%
}
\caption{Hallucination detection performance (F1, Precision, Recall, PR-AUC, and PRR@$k$) for \voxtral on the \fleurs S2TT test set, with $k{=}10\%$.}
\label{tab:s2tt_voxtral}
\end{table}

\paragraph{Feature combination is important for S2TT.}

S2TT performance benefits from combining multiple attention-based metrics. For \voxtral, using only \audioratio features reduces the F1-score from 0.37 to 0.35 and the PR-AUC from 0.44 to 0.38, compared to the Top 300 configuration. \qwen exhibits a similar trend, with drops of 0.05 in both F1-score and PR-AUC when comparing \audioratio to Top 150. These results support that the S2TT task relies on complementary signals captured by different attention metrics. Combining features is therefore particularly important for translation.

\subsection{Task-specific attention heads dominate.}
The observed lack of cross-task generalisation raises a fundamental question: does the logistic regression model rely on a universal set of attention heads, or is head selection inherently task-dependent? To investigate this, we compute the intersection of the Top 50 most informative attention heads for each metric across ASR and S2TT, as reported in Table~\ref{tab:attention_head_comp}. We select the Top 50 heads because they capture most of the discriminative signal, as shown by the scaling behaviour in Figure~\ref{fig:scaling}.

Across all metrics and both models, we observe limited overlap between selected heads. Combined with poor cross-task transfer, this suggests that hallucination-related attention features are largely task-specific. However, the low overlap may also reflect feature collinearity, in which redundant heads lead the model to select different yet functionally similar features across tasks.

\begin{table}[htbp] 
\centering 
\footnotesize 
\begin{tabular}{@{}lcc@{}} 
\toprule 
 & \multicolumn{2}{c}{Common Heads} \\ 
\cmidrule(lr){2-3} 
Metric & \qwen & \voxtral \\ 
\midrule 
\audioratio & 22\% & 18\% \\ 
\audioconsistency & 32\% & 26\% \\ 
\audioentropy & 10\% & 8\% \\ 
\textentropy & 14\% & 14\% \\ 
\bottomrule 
\end{tabular} 
\caption{Intersection of the top-50 most important attention heads for each metric for the ASR and S2TT tasks.} 
\label{tab:attention_head_comp} 
\end{table}

Among the four metrics, \audioconsistency shows the highest cross-task stability, followed by \audioratio, consistent with earlier results showing strong performance even with few heads. In contrast, both entropy-based metrics exhibit substantially lower consistency. This may be because entropy-based regressors assign high weights to heads with weak alignment to their nominal input modality. For example, \audioentropy may select heads that primarily attend to the auto-regressive text prefix rather than audio, making the resulting features more sensitive to task-specific noise.

\section{Conclusions}
We presented an attention-based approach to hallucination detection in SpeechLLMs. We introduced four audio-focused attention metrics, \audioratio, \audioconsistency, \audioentropy, and \textentropy, and used them to train lightweight logistic regression classifiers. We evaluated the approach on two SpeechLLMs across automatic speech recognition and speech-to-text translation tasks.

Our method outperforms all baselines on in-domain ASR, achieving improvements of up to +0.23 PR-AUC on \voxpopuli with \voxtral, with particularly strong gains on cleaner data. Reducing the feature set from over 1,000 attention heads to approximately 100 yields comparable in-domain performance while improving out-of-domain generalisation, indicating that larger feature sets are prone to overfitting. Effectiveness varies across models: \voxtral generalises well across settings, whereas \qwen exhibits weaker transfer to the noisier \callhome dataset.

Models trained on ASR data do not generalise to S2TT, highlighting that hallucination patterns and their associated attention signals are task-specific. However, training on S2TT data yields comparable improvements over baselines, demonstrating that the approach is effective when task-appropriate supervision is available. Future work could combine attention-based features with uncertainty estimation metrics to capture complementary signals, explore cross-model transfer, and extend the method to additional speech tasks such as speech summarisation or spoken dialogue systems.

\section{Limitations.}
Our automatic labelling strategy achieves high precision (0.979) but low recall (0.443), which means that many true hallucinations are excluded from training. Beyond threshold calibration, we do not perform direct human evaluation of the automatically labelled data. In addition, we focus exclusively on binary detection of severe hallucinations, rather than modelling finer-grained distinctions in hallucination type or severity.

Our approach does not generalise across tasks. Classifiers trained on ASR data do not generalise to S2TT, necessitating task-specific training data. Effectiveness is also model-dependent, with \voxtral exhibiting more robust cross-domain generalisation than \qwen. Moreover, our evaluation is limited to two SpeechLLMs and four languages, which constrains the scope of our conclusions.

Finally, although the proposed method is lightweight compared to alternatives such as resampling or ensemble-based approaches, extracting attention patterns introduces additional inference-time computation and memory overhead. While this overhead remains modest relative to full SpeechLLM inference, it may still be relevant in latency-sensitive deployment settings.
\clearpage

\bibliography{anthology_trimmed}

\appendix
\clearpage
\section{Logistic Regression}

All logistic regression models are trained using \texttt{scikit-learn}\footnote{\url{https://scikit-learn.org/stable/index.html}} \citep{scikit-learn}.

\begin{table}[h]
\centering
\begin{tabular}{lll}
\toprule
\textbf{Hyperparameter} & \textbf{L2} & \textbf{L1} \\
\midrule
Penalty & \texttt{L2} & \texttt{L1} \\
Max iterations & 5000 & 5000 \\
Class weights & 1:2 (positive) & 1:5 (positive) \\
C & 1 & 0.005 \\
Solver & \texttt{lbfgs} & \texttt{liblinear} \\
\bottomrule
\end{tabular}
\caption{Hyperparameters for L2 logistic regression and L1 logistic regression, the latter used for stable feature selection. These parameters were selected on the \voxpopuli validation set.}
\label{tab:hyperparams}
\end{table}

Table~\ref{tab:hyperparams} lists the hyperparameters used to train our hallucination detection models, selected on the \voxpopuli validation set. Unless otherwise stated, these parameters are used for all logistic regression models.

\begin{table}[htbp]
\centering
\begin{tabular}{@{}lc@{}}
\toprule
Metric & Features \\
\midrule
\audioratio & 39 \\
\audioconsistency & 36 \\
\audioentropy & 20 \\
\textentropy & 4 \\
\midrule
Total & 99 \\
\bottomrule
\end{tabular}
\caption{Distribution of stable features (threshold $\geq 0.8$, selected in 4 out of 5 folds).}
\label{tab:stable_features}
\end{table}

Table~\ref{tab:stable_features} summarises the stable features selected using L1 regularisation. These are the heads present in 4 out of 5 folds after training on the \voxpopuli training data. The distribution shows that features are selected across all heads, with \textentropy being a clear outlier, contributing only 4 features.

\section{S2TT Results}
\label{sec:s2tt_results}

Tables~\ref{tab:s2tt_performance_qwen} and~\ref{tab:s2tt_performance_voxtral} report chrF~\citep{popovic-2015-chrf} and COMET scores, computed using XCOMET-XL.

\begin{table}[htbp] 
\centering 
\begin{tabular}{@{}lcc@{}} 
\toprule 
Language Direction & chrF & COMET \\ 
\midrule 
En$\rightarrow$De & 54.19 & 87.06 \\ 
En$\rightarrow$Es & 47.42 & 82.26 \\ 
En$\rightarrow$Fr & 49.72 & 76.74 \\ 
De$\rightarrow$En & 60.29 & 90.33 \\ 
Es$\rightarrow$En & 54.43 & 87.20 \\ 
Fr$\rightarrow$En & 60.17 & 86.37 \\ 
\bottomrule 
\end{tabular} 
\caption{S2TT performance for \qwen on \fleurs.} 
\label{tab:s2tt_performance_qwen} 
\end{table}

The results mirror those in Tables~\ref{tab:hallucination_detection_qwen} and~\ref{tab:hallucination_detection_voxtral}, with \voxtral outperforming \qwen across both metrics and all datasets.

\begin{table}[htbp] 
\centering 
\begin{tabular}{@{}lcc@{}} 
\toprule 
Language Direction & chrF & COMET \\ 
\midrule 
En$\rightarrow$De & 57.73 & 90.65 \\ 
En$\rightarrow$Es & 51.09 & 88.69 \\ 
En$\rightarrow$Fr & 59.23 & 84.13 \\ 
De$\rightarrow$En & 64.90 & 94.42 \\ 
Es$\rightarrow$En & 59.11 & 92.07 \\ 
Fr$\rightarrow$En & 63.25 & 90.82 \\ 
\bottomrule 
\end{tabular} 
\caption{S2TT performance for \voxtral on \fleurs.} 
\label{tab:s2tt_performance_voxtral} 
\end{table}

\section{Semantic Hallucination Score}
\label{sec:shs}

\begin{figure*}[t]
    \centering
    \includegraphics[width=\textwidth]{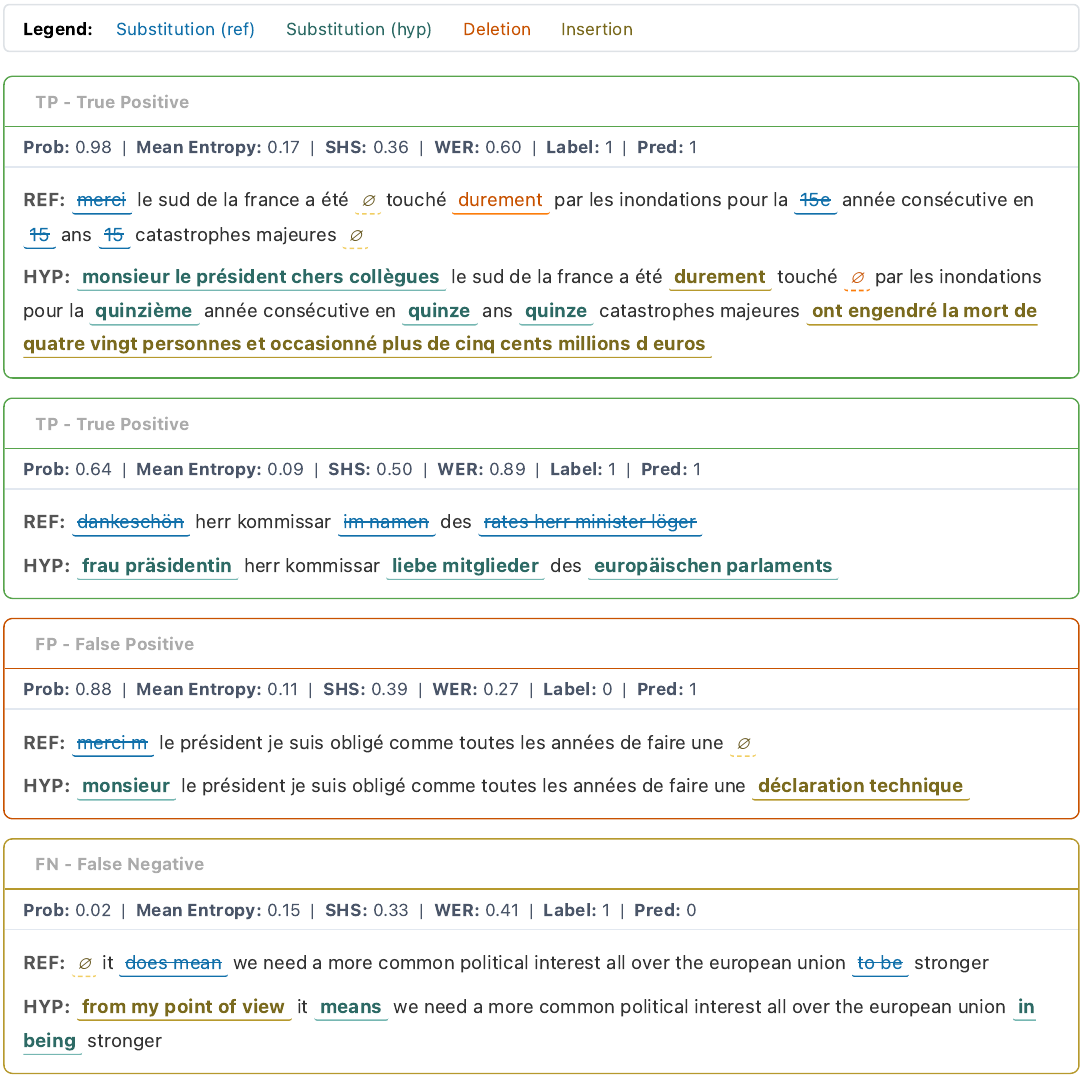}
    \caption{Examples of classifications using Top 75 with \voxtral outputs on the \voxpopuli test set.}
    \label{fig:examples}
\end{figure*}

A defining property of hallucinations is substantial semantic divergence between the gold standard transcription and the hypothesis. To capture this semantic change, we adopt the semantic hallucination score (SHS) from the SHALLOW ASR hallucination benchmark~\citep{koudounas2025hallucinationbenchmarkspeechfoundation}. SHS combines local and global semantic error metrics to capture both fine-grained and utterance-level inconsistencies between hypothesis and reference transcriptions.

Local semantic errors are computed using a multi-scale sliding window approach. Each hypothesis window is matched to reference windows via maximum cosine similarity of contextual embeddings, with higher weights assigned to smaller windows. This emphasises token-level distortions while remaining sensitive to phrase-level mismatches. Global semantic errors comprise two components. The first is semantic distance, computed as the inverse cosine similarity of sentence embeddings. The second is semantic coherence, which combines BERTScore with an entailment probability derived from a natural language inference model.

As our data spans multiple languages, we replace the original monolingual models with multilingual alternatives. We use \texttt{xlm-roberta-base}~\citep{conneau-etal-2020-unsupervised} for local embeddings, \texttt{paraphrase-multilingual-MiniLM}~\citep{reimers-gurevych-2019-sentence} for sentence embeddings, and \texttt{mDeBERTa-v3-base-xnli}~\citep{laurer_less_2022} for natural language inference.

\section{PRR Metric}
\label{sec:prr}

Equation~\ref{eq:prr} defines the predicted rejection ratio (PRR). For a corpus $D = \{(x_j, y_j)\}$, let $p_j = P(x_j, y_j^{*})$ denote the logistic regression model's predicted probability that hypothesis $y_j^{*}$ is a hallucination. The rejection curve plots the average quality $Q(y_j, y_j^{*})$ of the remaining samples after rejecting those with $p_j < \alpha$.

\begin{equation} 
\text{PRR} = \frac{\text{AUC}_{\text{prob}} - \text{AUC}_{\text{random}}}{\text{AUC}_{\text{oracle}} - \text{AUC}_{\text{random}}} 
\label{eq:prr}
\end{equation}

PRR measures how much average quality improves when rejecting samples based on predicted probability, relative to oracle rejection. A PRR of 1 indicates rejection in exact oracle order. We report PRR over the first 10\% of the rejection curve for \voxpopuli and the first 30\% for \callhome, reflecting the proportion of hallucinations in each dataset.

\section{\voxpopuli Examples}
\label{sec:examples}

Figure~\ref{fig:examples} presents four examples of \voxtral outputs classified using Top 75 on the \voxpopuli test set. The first two are true positives that \textsc{Mean Entropy} misclassifies as non-hallucinations. The first contains both substitutions and insertions, while the second exhibits hallucination through substitution alone. We also include one false positive and one false negative to illustrate typical failure modes. Both cases lie close to the decision boundary of our automatic labelling pipeline, highlighting the inherent difficulty of the task. Notably, \textsc{Mean Entropy} also misclassifies both examples.

\section{GPU Usage and AI Statement}

Inference and labelling are performed on eight \texttt{A100-40GB} GPUs, processing approximately 4.5 samples per second. A single experimental run covers approximately 57,000 ASR sentences and 21,000 S2TT sentences, requiring around 38.5 GPU hours. Over development and evaluation, we conducted approximately six complete iterations, totalling around 230 GPU hours. Including XCOMET scoring and SHS computation, we estimate an upper bound of approximately 300 GPU hours.

Code for this project was partially written with the assistance of an internal coding assistant. Internal AI tools were also used to assist with the language and presentation of the paper.

\end{document}